\documentclass[conference]{IEEEtran}
\IEEEoverridecommandlockouts
\newcommand*{\rom}[1]{\expandafter\@slowromancap\romannumeral #1@}

\usepackage{cite}
\usepackage{amsmath,amssymb,amsfonts}
\usepackage{algorithmic}
\usepackage{graphicx}
\usepackage{textcomp}
\usepackage{xcolor}
\usepackage{subcaption}
\usepackage{hyperref} 

\def\BibTeX{{\rm B\kern-.05em{\sc i\kern-.025em b}\kern-.08em
    T\kern-.1667em\lower.7ex\hbox{E}\kern-.125emX}}
\begin{document}

\title{Road Damage Detection And Classification In Smartphone Captured Images Using Mask R-CNN}

\author{\IEEEauthorblockN{Janpreet Singh* \thanks{*equal contribution}}
\IEEEauthorblockA{\textit{Data Science Division} \\
\textit{Atkins}\\
Bangalore, India \\
janpreetsingh.1@gmail.com
}
\and
\IEEEauthorblockN{Shashank Shekhar*$\dagger$ \thanks{$\dagger$corresponding author}}
\IEEEauthorblockA{\textit{Department Of Computational and Data Sciences} \\
\textit{Indian Institute Of Science}\\
Bangalore, India \\
shashankshek@iisc.ac.in}
}

\maketitle

\begin{abstract}
This paper summarizes the design, experiments and results of our solution to the Road Damage Detection and Classification Challenge held as part of the 2018 IEEE International Conference On Big Data Cup. Automatic detection and classification of damage in roads is an essential problem for multiple applications like maintenance and autonomous driving. We demonstrate that convolutional neural net based instance detection and classfication approaches can be used to solve this problem. In particular we show that Mask-RCNN, one of the state-of-the-art algorithms for object detection, localization and instance segmentation of natural images, can be used to perform this task in a fast manner with effective results. We achieve a mean F1 score of 0.528 at an IoU of 50\% on the task of detection and classification of different types of damages in real-world road images acquired using a smart-phone camera and our average inference time for each image is 0.105 seconds on an NVIDIA GeForce 1080Ti graphic card. The code and saved models for our approach can be found here : \href{https://github.com/sshkhr/BigDataCup18_Submission}
{\textit{https://github.com/sshkhr/BigDataCup18\_Submission}}\newline
\end{abstract}


\section{Introduction}
Automatic detection and classification of damage from road images has emerged as an important goal because road images provide basic information for several natural image based applications like autonomous driving. The task is challenging in two aspects. First, a robust damage detection and classification algorithm is required to localize individual damages over time under varying weather or lighting conditions. Second, the algorithm should be able to distinguish between overlapping damages of different types which is a very commonly observed phenomena in these damages.

Since the success of AlexNet \cite{krizhevsky2012imagenet} in the ImageNet Large Scale Visual Recognition Challenge \cite{ILSVRC15} algorithms based on Convolutional Neural Nets (CNNs) have become the de facto approach to computer vision problems and have led to significant advances in the state of the art for fundamental problems like image classification \cite{simonyan2014very} \cite{he2015delving}, object detection \cite{lin2017feature}\cite{ren2015faster} and semantic segmentation \cite{ronneberger2015u} among others.

We have used Mask R-CNN \cite{he2017mask} which is an object instance segmentation model alongside object detection and classification for the problem of road damage detection and classification. The rest of the paper is organised as follows : Section II describes the dataset used and evaluation protocol of the IEEE BigData 2018 Cup Challenge, Section III summarizes related work in the area and the Mask R-CNN architecture, Section IV discusses our implemetation, experiments and training approaches, Section V reports our results in terms of both classification and detection accuracy as well as inference time and finally Section VI draws conclusions from our work.

\section{Dataset And Evaluation Strategy}\label{eval}

The dataset for our experiments was taken from \cite{maedaroad} and consists of 9,053 labeled road damage images acquired from a smartphone camera. There are a total of 15,435 bounding boxes of damage annotated for the dataset which belong to 8 different classes. Each image in the dataset has a resolution of $600\times600$ pixels. 

During evaluation first a match is defined as:
\begin{itemize}
\item The predicted bounding box has the same class label as the ground truth bounding box.
\item The predicted bounding box has over 50\% Intersection over Union (IoU) in area with the ground truth bounding box. 
\end{itemize}
Then the evaluation of the match is done using the Mean $F_1$ Score metric. The $F_1$ score, commonly used in information retrieval, measures accuracy using the statistics of precision \textit{p} and recall \textit{r}. Precision is the ratio of true positives \textit{(tp)} to all predicted positives \textit{(tp + fp)} while recall is the ratio of true positives to all actual positives \textit{(tp + fn)}. The F1 score is given by:
\[
F_{1}=2 \times \frac{\text{p} \times \text{r}}{\text{p} + \text{r}} \ where \ \textrm{p} = \frac{\text{tp} }{\text{tp} + \text{fp}} \ and \ \textrm{r} = \frac{\text{tp} }{\text{tp} + \text{fn}}
\]
\vspace{1pt}

\begin{figure*}[htbp]
\centerline{\includegraphics[width=15cm,height=9cm,keepaspectratio]{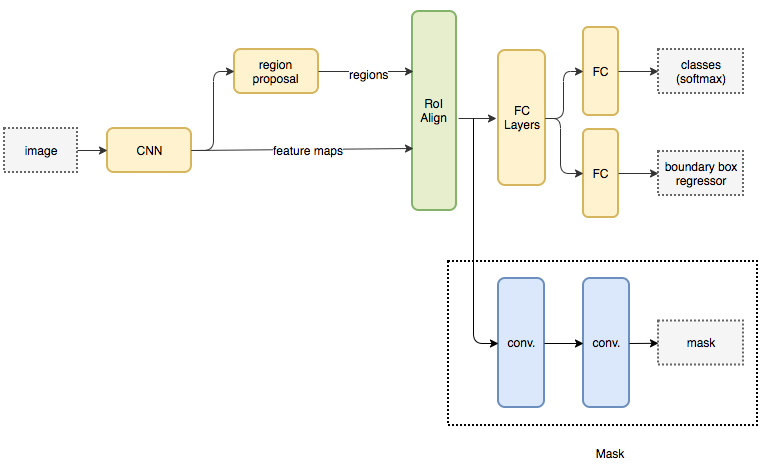}}
\caption{Mask R-CNN architecture}
\label{fig}
\end{figure*}

\section{Related Work}
Road damage detection and classification has been an active area of research for the vision and civil engineering community. Although there had been a lot of work \cite{doi:10.1111/mice.12042} \cite{doi:10.1111/j.1467-8667.2011.00716.x} on applying image processing approaches for the problem \cite{7533052} were the first to apply CNNs to road damage detection. Since then other works \cite{8476795} \cite{doi:10.1111/mice.12297} \cite{akarsu2016fast} have focused on using deep learning for crack detection in road images. 

However other than \cite{maedaroad} most of these works have been limited to detection of one particular type of damage or classifying images based on damage type. We now discuss Mask R-CNN which can be use to simultaneously detect and classify different types of road damage.

\subsection{Mask R-CNN}\label{AA}
Mask R-CNN is an extension of the Faster R-CNN model for object detection, localization and instance segmentation in natural images. This section summarizes the different components of Mask R-CNN and their use in the detection pipeline (see Fig. \ref{fig} for details of the architecture).  
\newline
\subsubsection{Backbone Network}
The backbone network for the Mask R-CNN is a standard CNN which is used to extract high level visual features from the entire image. ResNet networks of depth 50 and 101 are evaluated for the backbone network and shown to work well. 

The Feature Pyramid Network (FPN) \cite{lin2017feature} was introduced as a top-down pyramid architecture with takes high level features from first pyramid and passes to lower layers using lateral connections. This allows every level to have access to both low and high level visual features. It was demonstrated that using a ResNet-FPN as the backbone network for feature extraction with Mask R-CNN gives excellent gains in both accuracy and speed compared to simply using ReNet as backbone network.
\newline
\subsubsection{Region Proposal Network}
The Region Proposal Network (RPN) first introduced in Faster R-CNN is a fully convolutional network which takes in image features from the backbone network and proposes candidate object bounding boxes with their objectness score. This network replaced slower mechanisms for generating candidate bounding boxes like selective search \cite{uijlings2013selective} by simultaneously predicting K proposals from each sliding window location in the feature map generated by backbone network. These K proposals are parameterized relative to K \textit{anchor boxes} which are centred at the sliding window in question and associated with a scale and an aspect ratio.

The RPN generates an anchor class (foreground or background) and a bounding box refinement for each proposal. Top N refined bounding boxes which have the highest probability of being from foreground class are chosen. Since RPN proposals highly overlap with each other, the authors adopted Non-Maximum
Suppression (NMS) on the proposal regions based on their class scores in order to reduce redundancy. The IoU threshold for NMS was fixed at 0.7, which leaves about 2000 proposal regions per image.
\newline
\subsubsection{RoIAlign}
In order to predict pixel masks accurately Mask R-CNN requires the RoI features (which are small feature maps) to be well aligned
to accurately preserve the per-pixel spatial correspondence. For this purpose Mask R-CNN replaced the RoIPool layer in Faster R-CNN with an RoIAlign layer. The RoIAlign layer uses bi-linear interpolation to compute the exact values of the input features at four regularly sampled locations in each RoI bin, and then performs max or average pooling on the features. Since it doesn't use any quantization of features unlike RoIPool it manages to perform pixel-to-pixel alignment between network inputs and outputs.
\newline
\subsubsection{Network Head (Bounding Box Regression, Classification and Mask Prediction)}
Finally the RoIAligned features are passed to the network head which performs three parallel tasks of bounding box regression, classification and mask prediction. The classifier and regressor output the class labels and bounding box offsets collapsing the RoIAligned features into short output vectors by fully-connected (fc) layers. The mask branch predicts an $m \times m$ mask from each RoI using a Fully Convolutional Network (FCN) \cite{long2015fully}. This allows each layer in the mask branch to maintain the explicit $m \times m$ object spatial layout without collapsing it into a vector representation that lacks spatial dimensions.

\section{Methodology}

\subsection{Implementation Details}

For our work we modified Abdulla et al's \cite{matterport_maskrcnn_2017} excellent implementation of Mask R-CNN to perform our experiments. The implementation is done using Tensorflow \cite{abadi2016tensorflow} and uses RPN + ResNet101 as the backbone network. Due to the relatively small size of the dataset we use a Mask R-CNN model pre-trained on the MS-COCO dataset \cite{lin2014microsoft} as our starting off point. 

\subsection{Experiments}

Before training we initially resize the images to 512$\times$512 pixels. In order to alleviate the requirement of Mask R-CNN (and deep CNNs in general) for larger training data we performed data augmentation using horizontal flipping the images during training time. 

We experimented with the learning rate of the model in [0.00001, 0.0001, 0.001, 0.01].  While training the network we started training the initial layers in first two epochs and then trained the whole Mask R-CNN end to end along with learning rate annealing. After testing several groups of parameter combinations empirically, we get the best result using learning rate annealing by starting with 0.001 and decreasing it by a factor of 10 every 2 epochs. 

Finally we perform post processing by removing detection results of the same class with more than 0.85 Intersection-Over-Union among bounding boxes. In these cases the bounding box with the larger area was retained while the smaller bounding box discarded from our detection results.

We also tried a two-stage approach of first performing road segmentation using DeepLabV3 \cite{chen2017rethinking} and U-Net \cite{ronneberger2015u} trained using the CamVid dataset \cite{brostow2009semantic} and then running our end-to-end detector and classifier on the segmented road images. Our idea behind this approach was to remove spurious information in the data which will be fed to the second stage to allow the Mask R-CNN to focus only the road. However due to the large domain shift between CamVid and the target dataset this approach wasn't able to improve on our single-stage results.

\subsection{Training And Inference}

The training batch size was set to 4. All of our experiments were done on a NVIDIA GeForce GTX 1080Ti graphics card machine which has 11 GB DDR5X memory. 

While running our model for inference we kept the batch size of one image per batch in order to perform a fairer comparison across hardware devices. For comparison we have also run inference on NVIDIA GeForce GTX 1050Ti which has a smaller memory size of 4 GB DDR5 memory. The next section summarizes results of our experiments.

\section{Results}

\begin{figure}
\centering
\begin{subfigure}[b]{.45\linewidth}
\includegraphics[width=\linewidth]{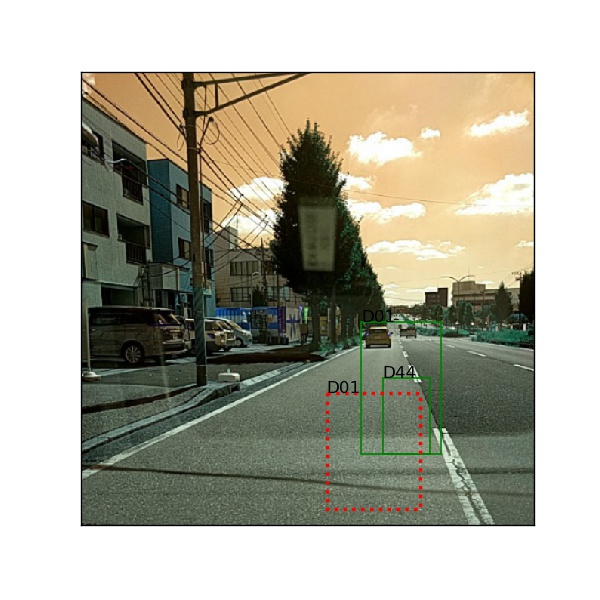}
\caption{Ichihara}\label{fig:ichi}
\end{subfigure}
\begin{subfigure}[b]{.45\linewidth}
\includegraphics[width=\linewidth]{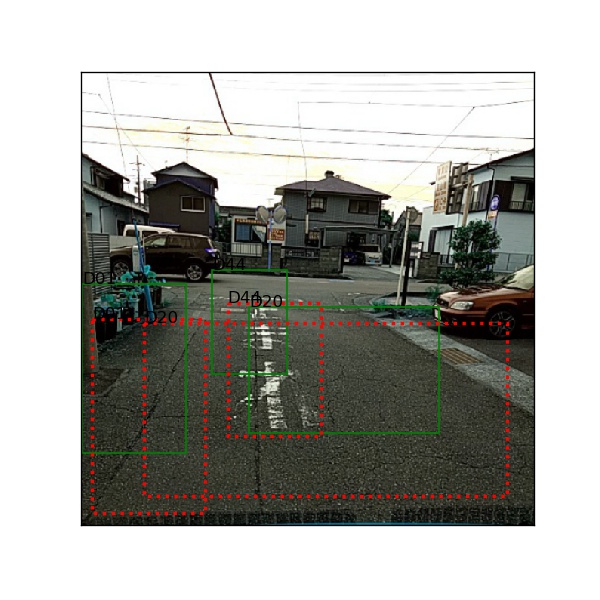}
\caption{Numazu}\label{fig:numa}
\end{subfigure}

\begin{subfigure}[b]{.45\linewidth}
\includegraphics[width=\linewidth]{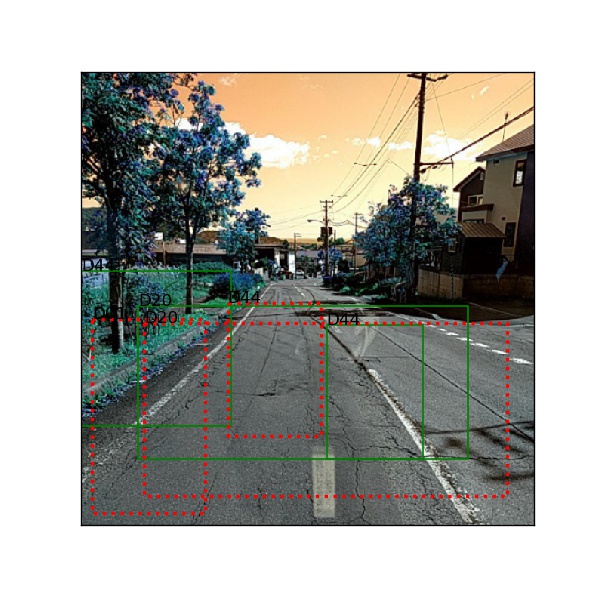}
\caption{Muroran}\label{fig:muro}
\end{subfigure}
\begin{subfigure}[b]{.45\linewidth}
\includegraphics[width=\linewidth]{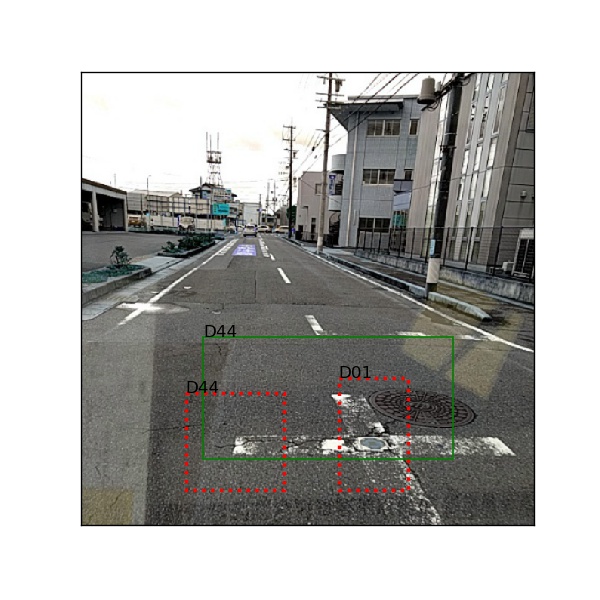}
\caption{Nagakute}\label{fig:naga}
\end{subfigure}
\caption{Detection Results from trained Mask R-CNN model for different areas under varying lighting conditions and viewpoints in the road damage dataset. The green bounding boxes reflect the ground truth and the red bounding boxes reflect the predictions of our model.}
\label{fig:animals}
\end{figure}

\paragraph{Detection And Classification} The evaluation was done using the platform provided by the organisers of the challenge. As described in Section  \ref{eval} earlier first those bounding boxes were selected whose class label matched with the ground truth and then those with a greater than 50\% Intersection-Over-Union were picked. Finally the Mean $F_1$ Score for these boxes was calculated which is summarised in Table \ref{tab1}

\begin{table}[htbp]
\caption{Detection Results}
\begin{center}
\begin{tabular}{|c|c|c|}
\hline
\textbf{Mean F-1 }&\multicolumn{2}{|c|}{\textbf{Dataset}} \\
\cline{2-3} 
\textbf{Score} & \textbf{\textit{Public Leaderboard}}& \textbf{\textit{Final Leaderboard}} \\
\hline
At 50 \% IoU $^{\mathrm{a}}$ & 0.528 & TBA \\
\hline
\multicolumn{3}{l}{$^{\mathrm{a}}$ Only correctly classified boxes were considered from all predictions}
\end{tabular}
\label{tab1}
\end{center}
\end{table}
\paragraph{Inference Speed} We calculated the average speed of inference for all 1813 test images for both GTX 1080Ti and GTX 1050Ti graphic cards. Table \ref{tab2} shows the time taken on average to run inference for our model.

\begin{table}[htbp]
\caption{Inference Speed}
\begin{center}
\begin{tabular}{|c|c|c|}
\hline
\textbf{Inference}&\multicolumn{2}{|c|}{\textbf{GPU Used}} \\
\cline{2-3} 
\textbf{Speed} & \textbf{\textit{GTX 1080Ti}}& \textbf{\textit{GTX 1050Ti}} \\
\hline
Avg time per image $^{\mathrm{a}}$ & 0.105 s & 0.285 s \\
\hline
Total time & 3 m 11 s & 8 m 38 s \\
\hline
\multicolumn{3}{l}{$^{\mathrm{a}}$ Images used were 512 $\times$ 512 pixels}
\end{tabular}
\label{tab2}
\end{center}
\end{table}

\section{Conclusions}

In this paper, we propose the use of convolution neural nets particularly Mask R-CNN for road damage detection and classification problem, which is formulated as an object detection and classification problem. In
our approach, Mask R-CNN is trained on real-world road images with bounding box information for the damage and its type. We use bounding box regression loss and classification loss together for the model to handle the detection and classification problem end-to-end. Comprehensive experimental results on the challenging road damage detection and classification dataset have demonstrated our approach of using Mask R-CNN for this problem performs as good as it's applications on natural images and common object classes.



\bibliography{references}{}

\begin{thebibliography}{10}

\bibitem{krizhevsky2012imagenet}
A.~Krizhevsky, I.~Sutskever, and G.~E. Hinton, ``Imagenet classification with
  deep convolutional neural networks,'' in {\em Advances in neural information
  processing systems}, pp.~1097--1105, 2012.

\bibitem{ILSVRC15}
O.~Russakovsky, J.~Deng, H.~Su, J.~Krause, S.~Satheesh, S.~Ma, Z.~Huang,
  A.~Karpathy, A.~Khosla, M.~Bernstein, A.~C. Berg, and L.~Fei-Fei, ``{ImageNet
  Large Scale Visual Recognition Challenge},'' {\em International Journal of
  Computer Vision (IJCV)}, vol.~115, no.~3, pp.~211--252, 2015.

\bibitem{simonyan2014very}
K.~Simonyan and A.~Zisserman, ``Very deep convolutional networks for
  large-scale image recognition,'' {\em arXiv preprint arXiv:1409.1556}, 2014.

\bibitem{he2015delving}
K.~He, X.~Zhang, S.~Ren, and J.~Sun, ``Delving deep into rectifiers: Surpassing
  human-level performance on imagenet classification,'' in {\em Proceedings of
  the IEEE international conference on computer vision}, pp.~1026--1034, 2015.

\bibitem{lin2017feature}
T.-Y. Lin, P.~Doll{\'a}r, R.~B. Girshick, K.~He, B.~Hariharan, and S.~J.
  Belongie, ``Feature pyramid networks for object detection.,'' in {\em CVPR},
  vol.~1, p.~4, 2017.

\bibitem{ren2015faster}
S.~Ren, K.~He, R.~Girshick, and J.~Sun, ``Faster r-cnn: Towards real-time
  object detection with region proposal networks,'' in {\em Advances in neural
  information processing systems}, pp.~91--99, 2015.

\bibitem{ronneberger2015u}
O.~Ronneberger, P.~Fischer, and T.~Brox, ``U-net: Convolutional networks for
  biomedical image segmentation,'' in {\em International Conference on Medical
  image computing and computer-assisted intervention}, pp.~234--241, Springer,
  2015.

\bibitem{he2017mask}
K.~He, G.~Gkioxari, P.~Doll{\'a}r, and R.~Girshick, ``Mask r-cnn,'' in {\em
  Computer Vision (ICCV), 2017 IEEE International Conference on},
  pp.~2980--2988, IEEE, 2017.

\bibitem{maedaroad}
H.~Maeda, Y.~Sekimoto, T.~Seto, T.~Kashiyama, and H.~Omata, ``Road damage
  detection and classification using deep neural networks with smartphone
  images,'' {\em Computer-Aided Civil and Infrastructure Engineering}.

\bibitem{doi:10.1111/mice.12042}
E.~Zalama, J.~Gómez-García-Bermejo, R.~Medina, and J.~Llamas, ``Road crack
  detection using visual features extracted by gabor filters,'' {\em
  Computer-Aided Civil and Infrastructure Engineering}, vol.~29, no.~5,
  pp.~342--358.

\bibitem{doi:10.1111/j.1467-8667.2011.00716.x}
T.~Nishikawa, J.~Yoshida, T.~Sugiyama, and Y.~Fujino, ``Concrete crack
  detection by multiple sequential image filtering,'' {\em Computer-Aided Civil
  and Infrastructure Engineering}, vol.~27, no.~1, pp.~29--47.

\bibitem{7533052}
L.~Zhang, F.~Yang, Y.~D. Zhang, and Y.~J. Zhu, ``Road crack detection using
  deep convolutional neural network,'' in {\em 2016 IEEE International
  Conference on Image Processing (ICIP)}, pp.~3708--3712, Sept 2016.

\bibitem{8476795}
V.~Pereira, S.~Tamura, S.~Hayamizu, and H.~Fukai, ``A deep learning-based
  approach for road pothole detection in timor leste,'' in {\em 2018 IEEE
  International Conference on Service Operations and Logistics, and Informatics
  (SOLI)}, pp.~279--284, July 2018.

\bibitem{doi:10.1111/mice.12297}
A.~Zhang, K.~C.~P. Wang, B.~Li, E.~Yang, X.~Dai, Y.~Peng, Y.~Fei, Y.~Liu, J.~Q.
  Li, and C.~Chen, ``Automated pixel-level pavement crack detection on 3d
  asphalt surfaces using a deep-learning network,'' {\em Computer-Aided Civil
  and Infrastructure Engineering}, vol.~32, no.~10, pp.~805--819.

\bibitem{akarsu2016fast}
B.~Akarsu, M.~KARAK{\"O}SE, K.~PARLAK, A.~Erhan, and A.~SARIMADEN, ``A fast and
  adaptive road defect detection approach using computer vision with real time
  implementation,'' {\em International Journal of Applied Mathematics,
  Electronics and Computers}, vol.~4, no.~Special Issue-1, pp.~290--295, 2016.

\bibitem{uijlings2013selective}
J.~R. Uijlings, K.~E. Van De~Sande, T.~Gevers, and A.~W. Smeulders, ``Selective
  search for object recognition,'' {\em International journal of computer
  vision}, vol.~104, no.~2, pp.~154--171, 2013.

\bibitem{long2015fully}
J.~Long, E.~Shelhamer, and T.~Darrell, ``Fully convolutional networks for
  semantic segmentation,'' in {\em Proceedings of the IEEE conference on
  computer vision and pattern recognition}, pp.~3431--3440, 2015.

\bibitem{matterport_maskrcnn_2017}
W.~Abdulla, ``Mask r-cnn for object detection and instance segmentation on
  keras and tensorflow.'' \url{https://github.com/matterport/Mask_RCNN}, 2017.

\bibitem{abadi2016tensorflow}
M.~Abadi, P.~Barham, J.~Chen, Z.~Chen, A.~Davis, J.~Dean, M.~Devin,
  S.~Ghemawat, G.~Irving, M.~Isard, {\em et~al.}, ``Tensorflow: a system for
  large-scale machine learning.,''

\bibitem{lin2014microsoft}
T.-Y. Lin, M.~Maire, S.~Belongie, J.~Hays, P.~Perona, D.~Ramanan,
  P.~Doll{\'a}r, and C.~L. Zitnick, ``Microsoft coco: Common objects in
  context,'' in {\em European conference on computer vision}, pp.~740--755,
  Springer, 2014.

\bibitem{chen2017rethinking}
L.-C. Chen, G.~Papandreou, F.~Schroff, and H.~Adam, ``Rethinking atrous
  convolution for semantic image segmentation,'' {\em arXiv preprint
  arXiv:1706.05587}, 2017.

\bibitem{brostow2009semantic}
G.~J. Brostow, J.~Fauqueur, and R.~Cipolla, ``Semantic object classes in video:
  A high-definition ground truth database,'' {\em Pattern Recognition Letters},
  vol.~30, no.~2, pp.~88--97, 2009.

\end{thebibliography}
\bibliographystyle{ieeetr}
\end{document}